\documentclass[sn-basic, iicol]{sn-jnl}%

\jyear{2021}%

\theoremstyle{thmstyleone}%
\theoremstyle{thmstyletwo}%
\theoremstyle{thmstylethree}%

\begin{document}

\title[CClusnet-Inseg]{Occlusion-Resistant Instance Segmentation of Piglets in Farrowing Pens Using Center Clustering Network}


\author[1,2]{\fnm{Endai} \sur{Huang}}\email{edhuang2-c@my.cityu.edu.hk}

\author[2]{\fnm{Axiu} \sur{Mao}}\email{max.mao@my.cityu.edu.hk}
\author[1]{\fnm{Junhui} \sur{Hou}}\email{jh.hou@cityu.edu.hk}
\author[1]{\fnm{Yongjian} \sur{Wu}}\email{yongjiawu5-c@my.cityu.edu.hk}
\author[1]{\fnm{Weitao} \sur{Xu}}\email{weitaoxu@cityu.edu.hk}
\author[3]{\fnm{Maria} \sur{Ceballos}}\email{mariacamila.ceballos@ucalgary.ca}
\author[4]{\fnm{Thomas} \sur{Parsons}}\email{thd@vet.upenn.edu}

\author*[2]{\fnm{Kai} \sur{Liu}}\email{kailiu@cityu.edu.hk} 

\affil[1]{\orgdiv{Department of Computer Science}, \orgname{City University of Hong Kong}, \orgaddress{\city{Hong Kong}, \country{China}}}

\affil*[2]{\orgdiv{Department of Infectious Diseases and Public Health}, \orgname{City University of Hong Kong}, \orgaddress{\city{Hong Kong}, \country{China}}}

\affil[3]{\orgdiv{Department of Production Animal Health, Faculty of Veterinary Medicine}, \orgname{University of Calgary}, \orgaddress{\city{Calgary}, \state{Alberta}, \country{Canada}}}

\affil[4]{\orgdiv{Swine Teaching and Research Center, School of Veterinary Medicine}, \orgname{University of Pennsylvania}, \orgaddress{\city{Kennett Square}, \state{Pennsylvania}, \country{USA}}}



\abstract{Computer vision enables the development of new approaches to monitor the behavior, health, and welfare of animals. Instance segmentation is a high-precision method in computer vision for detecting individual animals of interest. This method can be used for in-depth analysis of animals, such as examining their subtle interactive behaviors, from videos and images. However, existing deep-learning-based instance segmentation methods have been mostly developed based on public datasets, which largely omit heavy occlusion problems; therefore, these methods have limitations in real-world applications involving object occlusions, such as farrowing pen systems used on pig farms in which the farrowing crates often impede the sow and piglets. In this paper, we adapt a Center Clustering Network originally designed for counting to achieve instance segmentation, dubbed as CClusnet-Inseg. Specifically, CClusnet-Inseg uses each pixel to predict object centers and trace these centers to form masks based on clustering results, which consists of a network for segmentation and center offset vector map, Density-Based Spatial Clustering of Applications with Noise (DBSCAN) algorithm, Centers-to-Mask (C2M), and Remain-Centers-to-Mask (RC2M) algorithms. In all, 4,600 images were extracted from six videos collected from three closed and three half-open farrowing crates to train and validate our method. CClusnet-Inseg achieves a mean average precision (mAP) of 84.1 and outperforms other methods compared in this study. We conduct comprehensive ablation studies to demonstrate the advantages and effectiveness of core modules of our method. In addition, we apply CClusnet-Inseg to multi-object tracking for animal monitoring, and the predicted object center that is a conjunct output could serve as an occlusion-resistant representation of the location of an object.}





\maketitle



\section{Introduction}\label{sec1}

The behavior, health, and welfare of animals are attracting increasing attention, and an assessment of these factors often requires long-term monitoring. However, manual observation and recording over long periods is time-consuming and labor-intensive; therefore, an automated, sensor-based system is required. Owing to their low-cost operation and non-intensive requirements, computer vision systems have been widely used to monitor various animal species such as cattle \citep{Chen2021}, broiler chickens \citep{Liu2021}, goats \citep{Wang2018}, fish \citep{White2006}, and pigs \citep{Gan2021}. Although these computer vision tools capture a large number of videos and images, the analysis of these digital data remains a challenge. 

For the analysis of animals in videos or images, a common practice is to first detect individual objects in each image (e.g., representative pixels of each object and the corresponding location of the object) and to then extract other information such as the posture and movement of an object in a video. Three detection methods and the corresponding representations are usually used: object detection with bounding boxes, keypoint detection with keypoints, and instance segmentation with masks. In instance segmentation, each distinct object of interest in an image is delineated by using masks, and objects can be assigned to different classes with pixel-level accuracy, thus narrowing the gap between the large bounding boxes in object detection and the sparse keypoints in target representation; this method has considerable potential for use in animal monitoring \citep{brunger2020panoptic}. 

Instance segmentation has been used in several animal monitoring applications in which instance segmentation of individual animals is required. For example, using instance segmentation, pixels corresponding to a sow were identified so that its body length and withers height could be extracted to estimate its body weight \citep{shi2016approach}. Instance segmentation masks were used to generate eigenvectors, which were then used by a kernel extreme learning machine to identify pig mounting behavior \citep{li2019mounting}. Instance segmentation was used as a key preprocessing step in the selection of regions of a pig for automatic weight measurement of the pig \citep{He2021a}. Instance segmentation with multi-object tracking was used to reduce the variance in the predicted weight of pigs \citep{He2021}. Nevertheless, these applications limited to large-size pigs and with no heavy occlusion. Therefore, the application of instance segmentation in farrowing pen remains to be verified. 

Deep-learning-based methods have emerged as the predominant choice in instance segmentation due to their high accuracy and ability to automatically extract target features. For example, Mask R-CNN was a representative two-stage instance segmentation method that it generated region-of-interests (ROIs) in the first stage and then segmented these ROIs in the second stage to achieve instance segmentation \citep{he2017mask}. Generally, these two-stage instance segmentation methods are unable to obtain real-time speeds (30 FPS), e.g., PANet \citep{Liu2018} and Mask Scoring R-CNN \citep{Huang2019}. In contrast, YOLACT++ was a representative one-stage instance segmentation approach that it generated a set of prototype masks and then combined them using per-instance mask coefficients to achieve instance segmentation \citep{bolya2020yolact++}. Other one-stage instance segmentation method includes CenterMask \citep{Lee2020}, SOLO \citep{wang2020solo}, and PolarMask \citep{Xie2020}.

\begin{figure}[h]%
\centering
\includegraphics[width=0.48\textwidth]{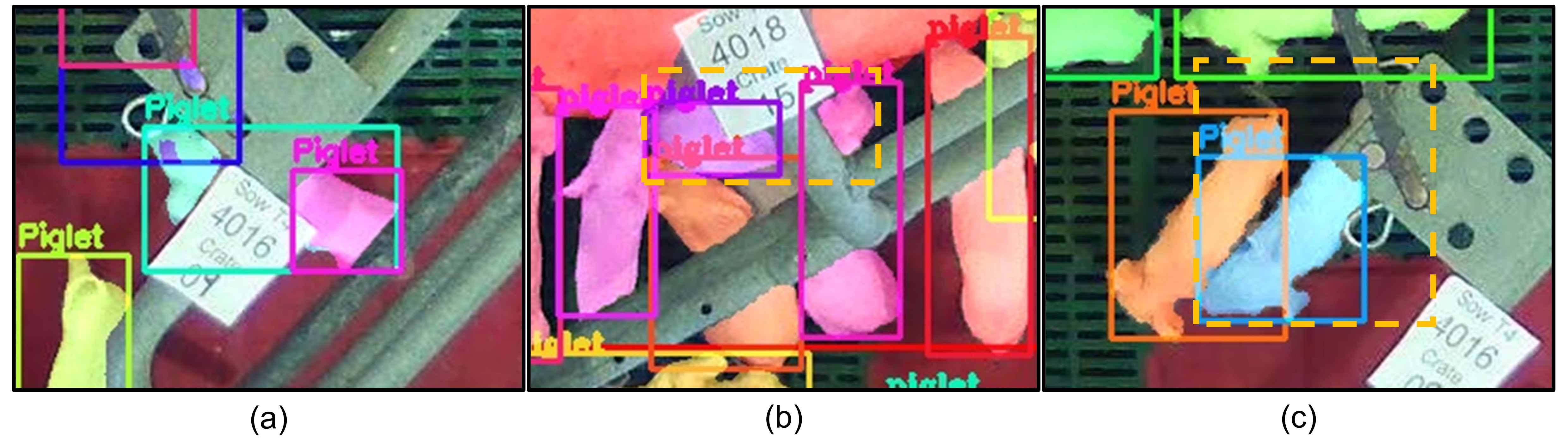}
\caption{Three common detection errors due to occlusion: (a) duplicate detection, (b) wrong association detection,  and (c) incomplete  detection.}
\label{fig1}
\end{figure}

However, most of the deep-learning-based methods for instance segmentation were developed using public datasets (e.g., Microsoft COCO; \citep{lin2014microsoft}). When these methods are used to monitor animals in commercial production settings, many scenarios may arise that are seldom featured in public datasets, such as frequently occurring occlusions. For instance, in pig farrowing pens, farrowing crates are widely used to restrict the movement of sows in order to reduce piglet mortality \citep{Moustsen2013, Johnson2009}. These farrowing crates introduce inevitable visual occlusions into the images and videos captured by computer vision systems. Such occlusions have non-negligible consequences for piglet due to piglets’ small size (e.g., only the head of a piglet may be visible while the other parts of the animal are occluded by farrowing crates). \cite{Huang2021} found that occlusions occurred in 97.7\% of the images collected from farrowing pens with farrowing crates. These unexpected occlusions may decrease the accuracy and even cause the failure of deep-learning-based methods \citep{huang2021capacity}. For example, there are three common types of detection errors due to occlusion: (1) Duplicate detection (Fig. \ref{fig1}a, by Blendmask \citep{chen2020blendmask}), where two or more duplicate detections overlap on a single object which is separated by occlusions; (2) Wrong association detection (Fig. \ref{fig1}b, by Mask R-CNN \citep{he2017mask}), where a disconnected part of a body (e.g., a head) is wrongly associated to another adjacent target; (c) Incomplete detection (Fig. \ref{fig1}c, by HoughNet \citep{samet2020houghnet}), where some parts of the body are missed in the detection due to occlusions.
Therefore, it is essential to develop a deep-learning-based instance segmentation method that can address occlusion problems. 

This study aims to develop a novel occlusion-resistant instance segmentation method for piglets in farrowing pens with frequently occurring occlusions. We propose CClusnet-Inseg, a Center Clustering Network for instance segmentation, based on a previous object-counting model called Center Clustering Network  (CClusnet-Count; \citep{Huang2021}). CClusnet-Count is designed for object counting, which predicts scatter object centers, and ends with the group number determined by the Mean-shift algorithm. We adapt this model to achieve instance segmentation. Our main contributions are summarized as follows:
\begin{enumerate}[1.]
\item We adapt a Center Clustering Network to a new framework for instance segmentation (CClusnet-Inseg) in farrowing pens. We change the original Mean-shift algorithm to a faster Density-Based Spatial Clustering of Applications with Noise (DBSCAN) algorithm, add a Centers-to-Mask (C2M) step, and further add a Remain-Centers-to-Mask (RC2M) step to fully use every pixel to form masks. CClusnet-Inseg is an anchor-free method with a high utilization of visible pixels in an image, and thus could manage heavy occlusion problems and outperform other instance segmentation methods.

\item We predict a unique object center along with instance segmentation for piglets. This predicted object center can be an occlusion-resistant representation of the object's location even when an object is largely occluded. 

\item We apply our method to multi-object tracking as a potential application for animal monitoring, and generate further animal characterizations including animal movement, trajectory, average speed, body pixel size, space usage, and spatial distribution. We demonstrate the limit of general instance segmentation and the necessity of our occlusion-resistant object center for animal monitoring under occlusion.
\end{enumerate}

\section{Materials and methods}\label{sec2}
\subsection{Data}
\begin{figure}[h]%
\centering
\includegraphics[width=0.48\textwidth]{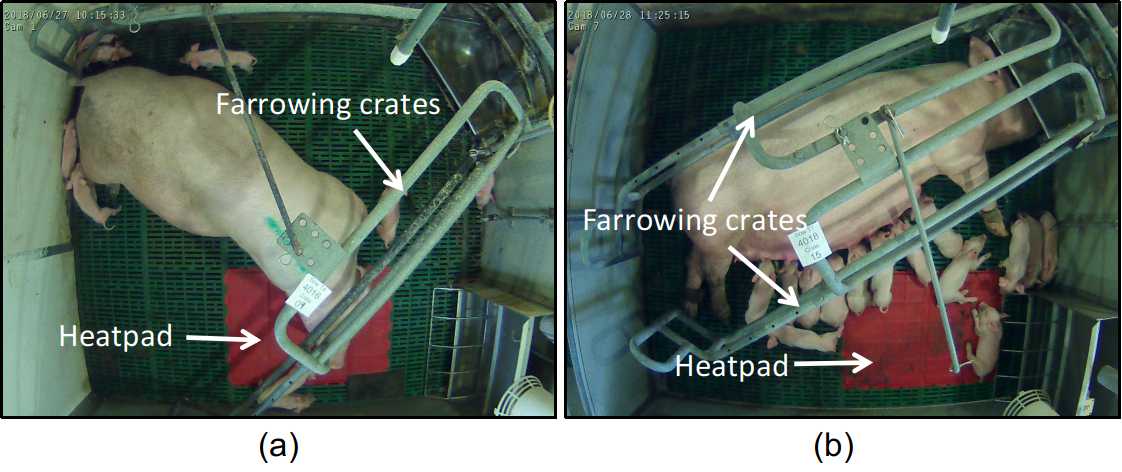}
\caption{Farrowing pens with (a) half-open and (b) fully closed farrowing crates}\label{fig2}
\end{figure}

The data was from a previously published study \citep{Ceballos2020} conducted at the Swine Teaching and Research Center, University of Pennsylvania School of Veterinary Medicine, United States, where 39 lactating sows and their offspring (Line 241; DNA Genetics, Columbus, NE) were initially used. All animal procedures in the previous study were approved by the University of Pennsylvania’s Institutional Animal Care and Use Committee (Protocol \#804656). Six top-view videos collected from 6:00 to 18:00 in June 2018 were selected for the development of the method in the current study (Tables~\ref{tab1}). The six videos, with a frame rate of 7 FPS and a resolution of 1024×768 pixels, were captured by overhead cameras located 2 m above the farrowing crates. Six datasets, containing 4,600 images in total, were extracted from the six videos and labeled for piglet centers, individual piglet masks, and sow masks \citep{Huang2021}. Datasets 1, 2, and 3 contained data from pens with half-open farrowing crates, and Datasets 4, 5, and 6 contained data from pens with closed crates (Fig. \ref{fig2}). Each video included one sow, and the number of visible piglets varied from 7 to 17 (an average of 11.4). 

Two types of partial occlusion on piglets were defined: body-separated occlusion (BSO) and part-missing occlusion (PMO) \citep{Huang2021}. In BSO, the occlusions (e.g., occlusions from farrowing crates) separated a target into two or more parts, whereas in PMO, a terminal part of a target (e.g., the head of a piglet) was occluded. The statistical summary showed that on an average, 3.7 piglets were affected by BSO, 4.2 piglets were affected by PMO, and 1.7 piglets were affected by both BSO and PMO per image. BSO occurred in 95.4\% of the images, and PMO occurred in 92.3\% of the images. Piglets were completely occluded (e.g., occluded by the sow, sow feeder, and farrowing crates) in 19.1\% of the images. Overall, 98.7\% of the images had at least one type of partial occlusion (BSO or PMO).

\begin{sidewaystable}
\sidewaystablefn%
\begin{center}
\begin{minipage}{\textheight}
\caption{Statistical summary for the datasets}\label{tab1}
\begin{tabular*}{\textwidth}{@{\extracolsep{\fill}}lccccccc@{\extracolsep{\fill}}}
\toprule
Description\footnotemark[1] & Dataset 1\footnotemark[2] & Dataset 2 & Dataset 3 & Dataset 4 & Dataset 5\footnotemark[3] & Dataset 6 & Total \\ 
\midrule
Number of days after farrowing & 5 & 5 & 5 & 4 & 5 & 5 & NA \\
Farrowing crates & Half-open & Half-open & Half-open & Closed & Closed & Closed & NA \\
Images extracted & 1,100 & 1,000 & 1,000 & 500 & 500 & 500 & 4,600 \\
Visible piglets\footnotemark[4] & 7–13 (8.9) & 7–9 (8.9) & 7–13 (12.6) & 10–17 (16.0) & 11–17 (15.6) & 9–11 (10.9) & 7–17 (11.4) \\
Piglets affected by BSO\footnotemark[4] & 0–6 (2.0) & 0–7 (2.9) & 0–9 (4.4) & 0–12 (4.2) & 0–13 (6.4) & 0–10 (4.0) & 0–13 (3.7) \\
Piglets affected by PMO\footnotemark[4] & 0–5 (1.4) & 0–7 (2.3) & 0–11 (5.7) & 2–16 (8.1) & 2–13 (7.2) & 0–11 (4.4) & 0–13 (4.2) \\
Piglets affected by both BSO and PMO\footnotemark[4] & 0–4 (0.8) & 0–5 (0.9) & 0–7 (1.9) & 0–9 (2.8) & 0–8 (3.3) & 0–9 (2.5) & 0–9 (1.7) \\
Images with BSO & 87.0\% & 98.3\% & 99.4\% & 95.4\% & 99.4\% & 96.2\% & 95.4\% \\
Images with PMO & 78.0\% & 89.7\% & 99.7\% & 100.0\% & 100\% & 98.8\% & 92.3\% \\
Images with complete occlusion & 9.6\% & 8.3\% & 24.7\% & 52.4\% & 29.6\% & 6.2\% & 19.1\% \\
Images with partial occlusion (BSO or PMO) & 94.8\% & 99.7\% & 100.0\% & 100.0\% & 100.0\% & 99.8\% & 98.7\% \\ 
\bottomrule
\end{tabular*}
\footnotetext[1]{BSO, body-separated occlusion; PMO, part-missing occlusion; NA, not available.}
\footnotetext[2]{Four piglets were removed from the pen due to low vitality or death.}
\footnotetext[3]{One piglet was removed from the pen due to low vitality or death.}
\footnotetext[4]{Per image, values in parentheses represent the average.}
\end{minipage}
\end{center}
\end{sidewaystable}

\subsection{CClusnet-Inseg}
\begin{figure*}[h]%
\centering
\includegraphics[width=0.95\textwidth]{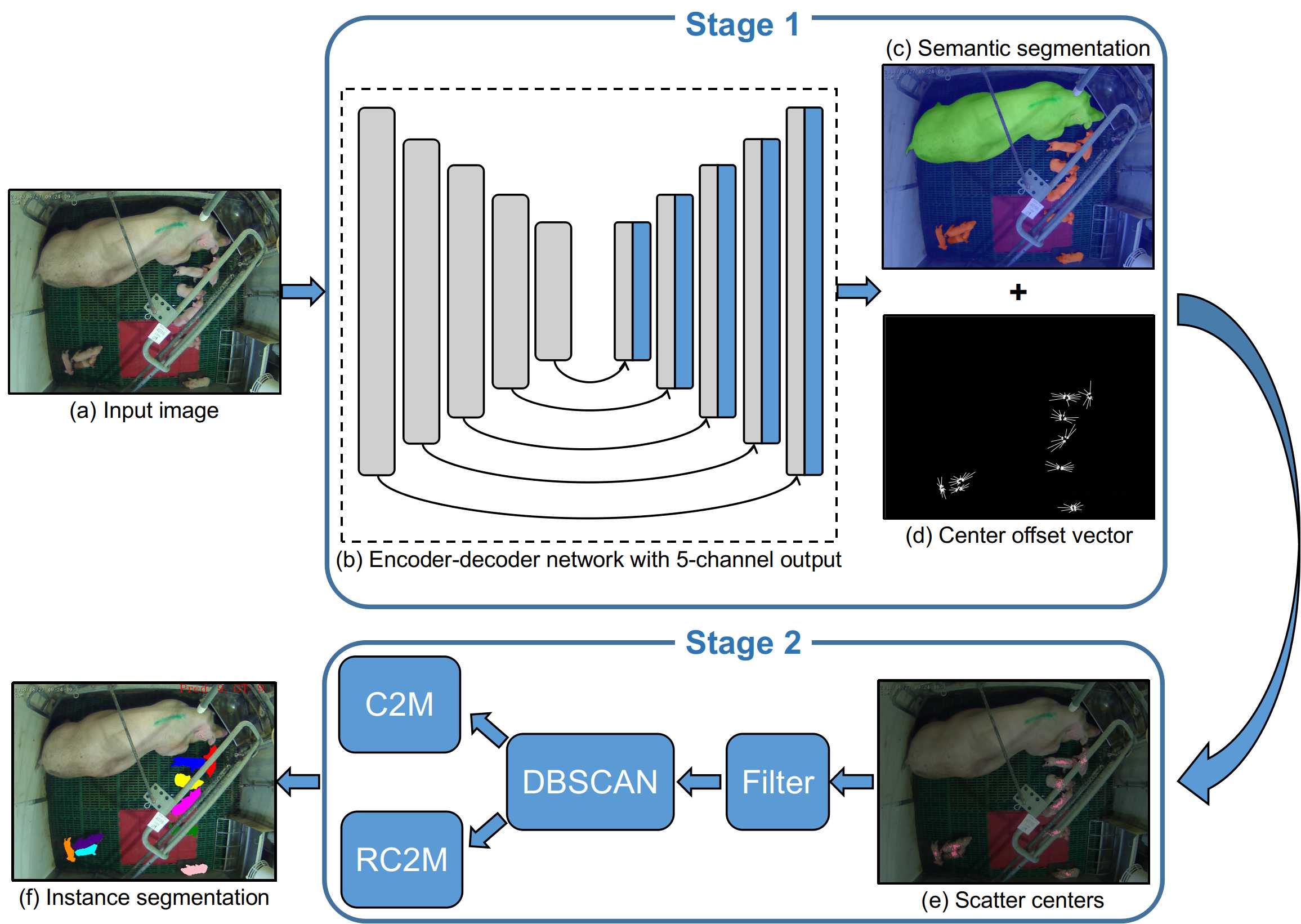}
\caption{The CClusnet-Inseg framework}\label{fig3}
\end{figure*}

\textbf{Overview.} The flow of CClusnet-Inseg is as follows: First, we feed an input image to an encoder–decoder network (Figs.  \ref{fig3}b and  \ref{fig4}); the network generates a semantic segmentation map (Fig.  \ref{fig3}c) and a center offset vector map (Fig.  \ref{fig3}d) as output. For each pixel belonging to a piglet, a center point is predicted using its corresponding offset vector, and thus we obtain scatter center points (Fig.  \ref{fig3}e). In the second stage, we filter these predicted scatter center points to eliminate outliers (Eq. 6 in \citep{Huang2021}) and then we cluster them into different groups through the DBSCAN algorithm \citep{ester1996density}. After the clustering, a C2M step traces these clustered center points back to their original pixels, and original pixels whose center points are in the same group form a mask of an instance. As some centers are filtered or clustered as noise in the DBSCAN algorithm, we perform a RC2M step to further cluster these centers and trace them back to their original pixels in the same manner as in the C2M step. 

\subsubsection{Network structure and loss function}
\begin{figure*}[h]%
\centering
\includegraphics[width=0.9\textwidth]{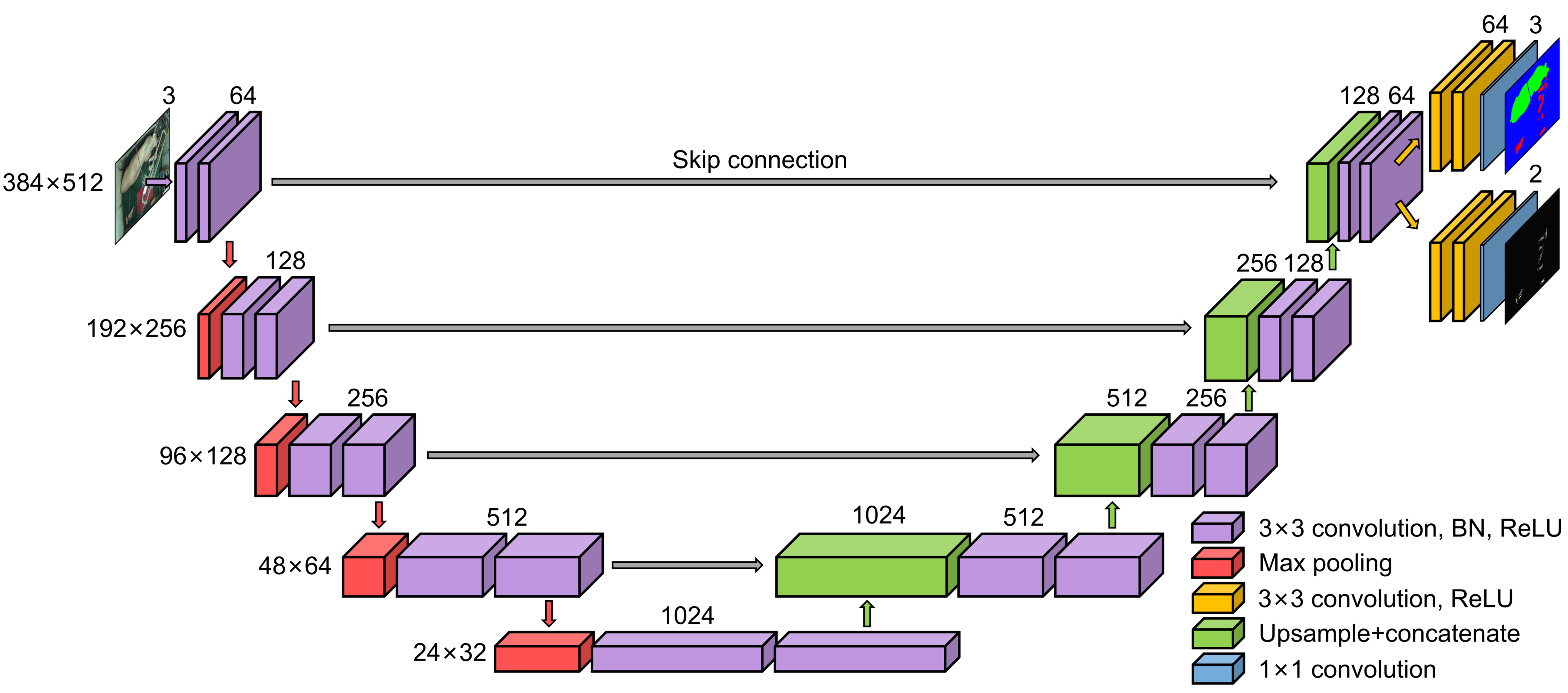}
\caption{Network architecture of CClusnet-Inseg}\label{fig4}
\end{figure*}

Our encoder–decoder network (Fig. \ref{fig3}b) structure, which is adapted from CClusnet-Count \citep{Huang2021}, is shown in Fig. \ref{fig4}. The input image is resized to $512 \times 384$ pixels and then fed to the network. Two separate heads, with two convolutional layers for semantic segmentation and four convolutional layers for the generation of the center offset vector, are attached at the end of the network. For the semantic segmentation task, the focal loss \citep{lin2017focal} $L_1$ is defined as

\begin{equation}
L_{1}=\frac{1}{N} \sum_{x}-\alpha_{t} \cdot\left(1-\hat{p}_{t}^{(x)}\right)^{\gamma} \cdot \log \left(\hat{p}_{t}^{(x)}\right)
\end{equation}
where
\begin{equation}
\hat{p}_{t}^{(x)}=\hat{p}^{(x)} \cdot p^{(x)}+\left(1-\hat{p}^{(x)}\right) \cdot\left(1-p^{(x)}\right),
\end{equation}
and $N$ is the number of pixels. $\alpha_{t}$ is the class weight for different classes, and $\hat{p}^{(x)}$ and $p^{(x)}$ are the predicted and true class probability vectors, respectively, for each pixel $x . \gamma$ is a focusing parameter that forces the model to focus on difficult examples.For the center offset vector generation task, the loss function $L_{2}$ is defined as
\begin{equation}
L_{2}=\frac{1}{N} \sum_{x} \operatorname{Mask}_{p l} \odot\|D(x)-\widehat{D}(x)\|_{2}^{2},
\end{equation}
where $D(x)$ and $\widehat{D}(x)$ are the ground truth and predicted displacement matrices, respectively, and $M a s k_{p l}$ is the binary mask of the piglet based on the semantic segmentation map, where 1 indicates that a pixel belongs to the piglet, and 0 indicates that it does not. The operation $\odot$ is the Hadamard product-the element-wise product of two matrices.

The final loss is
\begin{equation}
L=\lambda L_{1}+(1-\lambda) L_{2},\label{eq4}
\end{equation}
where $\lambda$ and $1-\lambda$ are the weights of $L_{1}$ and $L_{2}$, respectively. The main difference between the network structures of CClusnet-Inseg and CClusnet-Count is that CClusnet-Inseg consists of two separate convolutional layers in two heads, as these heads perform different functions.

\subsubsection{Centers-to-Mask (C2M) and Remain-Centers-to-Mask (RC2M)}
After the semantic segmentation map and center offset map are produced in the stage 1 (Fig.  \ref{fig3}), the scatter center points are generated as
\begin{equation}
C=\{P_i+O_i\}_{i=1}^N,
\end{equation}
where $P$ is the pixel set belonging to piglets in the semantic segmentation map and $O$ is the corresponding offset vector set in the center offset map. The DBSCAN algorithm is applied to these scatter center points to cluster them into different groups, and then the C2M step traces the clustered center points back to their original pixels in order to achieve instance segmentation. The number of groups determined by the DBSCAN algorithm is denoted as $M$. Each group is represented as
\begin{equation}
G_m=\{C_i \in C \mid  GID(P_i)=m\}_{i=1}^N, m=0,1, \cdots, M,
\end{equation}
where $GID()$ is a function that maps a piglet pixel to its group ID. The group ID ranges from 0 to $M$, where 0 means the group of points that are clustered as noise or filtered before the clustering algorithm.

The image consists of $M$ instances of piglets, which can be represented by $M$ binary mask matrices $\{PM_1, PM_2, \cdots, PM_M\}$ as:
\begin{equation}
PM_{m}^{(p)}=\begin{cases} 
1, & \mbox{if }GID(p)=m \\
0, & \mbox{otherwise}
\end{cases}, m=1,2, \cdots, M.
\end{equation}
where $p$ represents each pixel in the mask. As each image consisted of only one sow, the segmentation results for a sow could be directly used as instance segmentation results:
\begin{equation}
    SM^{(p)}=\left\{\begin{array}{l}1, \text { if } p \text { belongs to sow } \\ 0, \text { otherwise }\end{array}\right.
\end{equation}

The above step is called C2M. However, as the DBSCAN algorithm clusters some center points as noise and some centers are filtered before the clustering algorithm, some center points remain without a valid group. For each existing group, the average of the points in each group is used as its cluster center:
\begin{equation}
    C_m'= Average \left(G_{m}\right), m=1,2, \cdots, M,
\end{equation}
where $Average()$ calculates the average location of the center points in a group. The remaining unclassified center points in the group $G_0$ can be updated using the nearest cluster centers as follows:
\begin{equation}
    GID(P_i)=\underset{m}{\operatorname{argmin}}\left(distance \left(C_i, C_{m}'\right)\right), P_i \in G_{0}
\end{equation}

Therefore, we assign the group ID of the nearest cluster center to these unclassified center points. All of these unclassified center points are assigned a valid group ID. Then, the C2M step described above is performed to update the instance segmentation results. The step that uses the remaining unclassified centers to update the instance segmentation is called RC2M. The RC2M step does not affect the clustering but only the instance segmentation results.

\subsection{Performance evaluation and comparison}
The mean average precision (mAP), the most commonly used metric for instance segmentation, is used as the performance metric. To demonstrate and evaluate the performance of CClusnet-Inseg, we 1) perform an ablation study to determine the contributions and effects of each module—DBSCAN (compared with mean-shift as adopted in CClusnet-Count) and RC2M, and 2) compare it with six state-of-the-art instance segmentation methods—Mask R-CNN \citep{he2017mask}, SOLOv2  \citep{wang2020solov2}, Blendmask \citep{chen2020blendmask}, CondInst  \citep{tian2020conditional},  YOLACT++ \citep{bolya2020yolact++}, and HoughNet  \citep{samet2020houghnet}. The same training set and test set are used in all of the evaluations. 

\subsection{Applications of the method}
To demonstrate the potential applications of our method in animal monitoring, we apply the instance segmentation of piglets directly in the implementation of multi-object tracking in farrowing pens, which provides a solid foundation for subsequent in-depth animal monitoring tasks such as the characterization of the spatiotemporal distribution patterns of piglets. We select a half-minute video clip, from the video used in the generation of Dataset 1 and in which piglets and the sow are active, as a sample video. CClusnet-Inseg detects each frame in the video clip and provides the instance segmentation results as output. For pairing objects between two consecutive frames, the Intersection-over-Union (IoU) is calculated and used as the single metric to determine whether two instances are the same. We determine the object pair that had the maximum IoU among unpaired objects between two consecutive frames and assign the same ID to these objects. Then, we repeat the process for the remaining unpaired objects until no object pairs can be found. The unpaired new object in the current frame is assigned a new ID, and the unpaired old object in the previous frame is removed. 

From the multi-object tracking based on instance segmentation, we further extract six types of information—individual animal movement (accumulated movement in the video), trajectory, average speed, body pixel size, spatial usage, and spatial distribution. When generating animal movement, trajectory, and average speed, the predicted object center—a conjunct output of CClusnet-Inseg—is used to represent the location of an object. This predicted object center is valid even when an object is partially occluded. When calculating the body pixel size, the mean of the top five maximum mask areas of an object is used to exclude the case in which an object is partially occluded. To represent the spatial distribution, we generate a heat map by accumulating all of the pixels that an object had occupied in the video over its entire duration. 

\subsection{Experimental setup}
All of the images from Datasets $1-6$ were used; they were randomly divided such that $60 \%$ of the images were used as a training set, $20 \%$ as a validation set, and $20 \%$ as a test set. Thus, a training set of 2,760 images was constructed to train the model, which was validated and tested on a validation set and a test set that contained 920 images each. Image augmentations, such as rotation, flips, Gaussian blur, and pseudo-occlusion generation, were applied simultaneously to the training set during the training. The optimizer Adam \citep{Kingma2014} with a learning rate of $1 \times 10^{-4}$ was chosen for training. The weight $\lambda$ was set to $0.5$ (Eq. \ref{eq4}). The epoch was set to 300 iterations, and the batch size was set to 8. The three hyperparameters-threshold $t$ (Eq. 6 in \cite{Huang2021}), radius $\epsilon$, and the minimum number of points in DBSCAN-were set to 20, 2.5, and 50, respectively, based on the best results from the validation set. All experiments were conducted using Pytorch on a single NVIDIA GeForce RTX 3090 GPU.

\section{Results and discussion}
\subsection{Ablation study}
To demonstrate the improvement and influence of each extension (i.e., DBSCAN and RC2M), an ablation study was conducted (Table \ref{tab2}). The results showed that the accuracy of the model using DBSCAN (82.8 mAP in Model 2) was improved than that of the model using mean-shift (81.1 mAP in Model 1), whereas the clustering speed of the model using DBSCAN (0.152 s/image) was approximately 20 times as fast as that of the model using mean-shift (2.846 s/image), leading to a faster total inference speed in the model using DBSCAN (0.229 s/image as against 2.939 s/image in the model using mean-shift). This difference in speed was mainly due to the algorithm itself, i.e., the time complexity of the DBSCAN algorithm is $O(n \log n)$, which is less than that of the mean-shift algorithm, which is $O\left(n^{2}\right)$. The RC2M improved the instance segmentation accuracy by 1.3 mAP (an improvement from an mAP of 82.8 in Model 2 to 84.1 in Model 3) because it reused the unclassified center points. However, this step resulted in additional computations, and therefore, the inference time increased slightly by $2.2\%$. Thus, the use of the RC2M involves a trade-off between speed and accuracy. Finally, when all these extensions were included, CClusnet-Inseg achieved an mAP of 84.1. Therefore, the inclusion of all extensions was used as the default setting for our method in the subsequent experiments. 

\begin{table*}[h]
\begin{center}
\begin{minipage}{\textwidth}
\caption{Results from ablation study. The speed was calculated on average speed for the test set (920 images)}\label{tab2}
\begin{tabular*}{\textwidth}{@{\extracolsep{\fill}}lcccc@{\extracolsep{\fill}}}
\toprule
Model (clustering algorithm)  & RC2M & mAP & Clustering Speed & Total Speed \\ 
\midrule
1. CClusnet-Inseg (mean-shift)  & × & 81.1 & 2.846 s/image & 2.939 s/image \\
2. CClusnet-Inseg (DBSCAN)  & × & 82.8 & 0.152 s/image & 0.229 s/image \\
3. CClusnet-Inseg (DBSCAN) & \checkmark & 84.1 & 0.151 s/image & 0.234 s/image \\
\botrule
\end{tabular*}
\end{minipage}
\end{center}
\end{table*}

\subsection{Comparison with other methods}
The results of the performance comparison of our method with other instance segmentation approaches are shown in Table  \ref{tab3}. Houghnet, Mask R-CNN, SOLOv2, Blendmask, CondInst, and YOLACT++ achieved an mAP of 41.1, 74.4, 76.6, 79.8, 80.6, 80.9, respectively. CClusnet-Inseg outperformed these methods by achieving an mAP of 84.1.

\begin{table}[h]
\begin{center}
\begin{minipage}{\textwidth}
\caption{Comparison of mean average precision (mAP) \protect\\for different instance segmentation methods}\label{tab3}%
\begin{tabular}{@{}lcc@{}}
\toprule
Method & mAP & Speed \\ 
\midrule
HoughNet \citep{samet2020houghnet} & 41.1 & 0.10 s/image\\
Mask R-CNN \citep{he2017mask} & 74.4 & 0.06 s/image \\
SOLOv2 \citep{wang2020solov2} & 76.6 & 0.15 s/image \\
Blendmask \citep{chen2020blendmask} & 79.8 & 0.07 s/image \\
CondInst \citep{tian2020conditional} & 80.6 & 0.09 s/image \\
YOLACT++ \citep{bolya2020yolact++} & 80.9 & 0.08 s/image \\
CClusnet-Inseg (Ours) & 84.1 & 0.23 s/image \\
\botrule
\end{tabular}
\end{minipage}
\end{center}
\end{table}

\begin{figure*}[h]%
\centering
\includegraphics[width=0.98\textwidth]{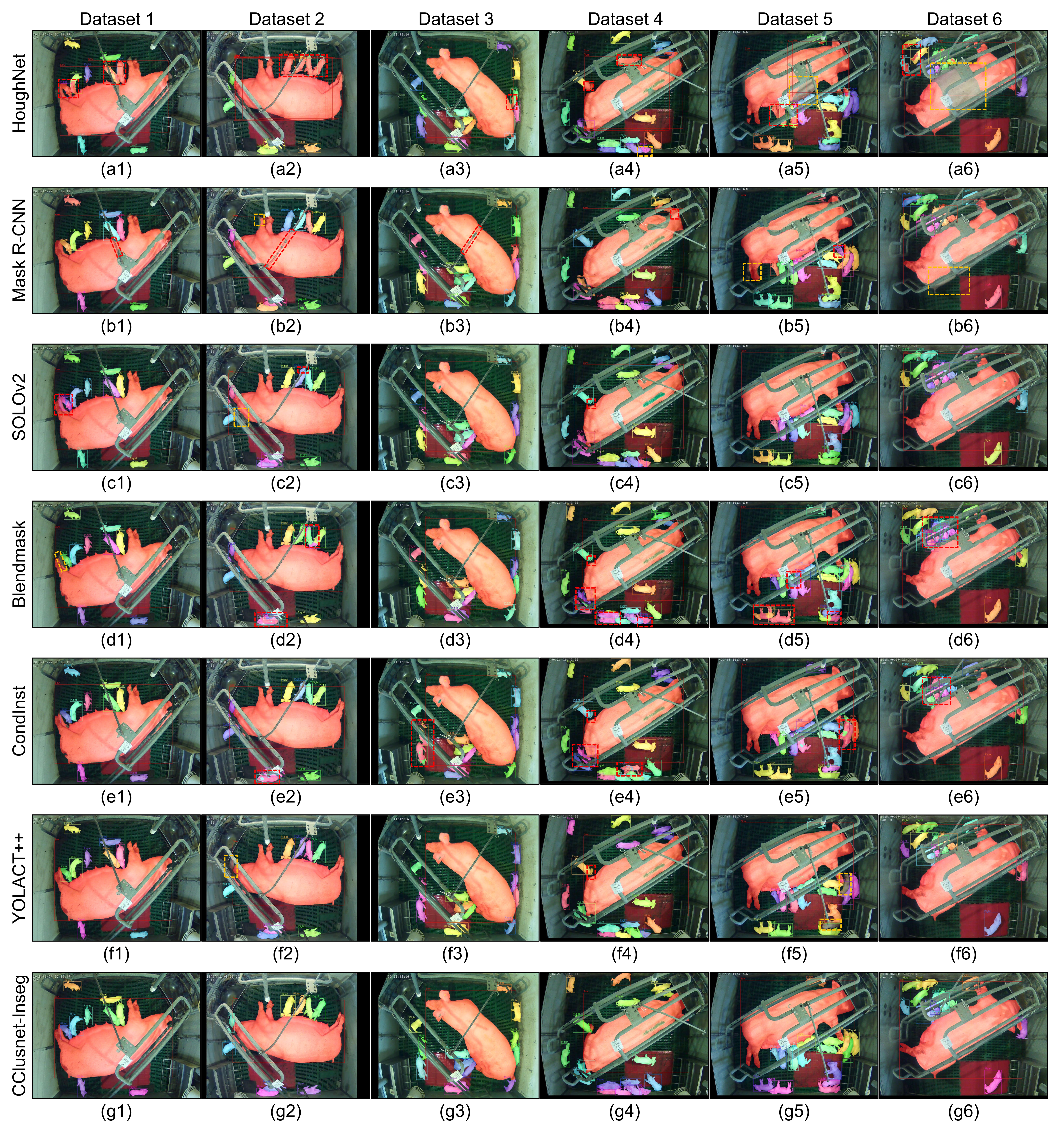}
\caption{Visualization of the results of the seven methods for each dataset. A significant misdetection is indicated by an orange dotted rectangle, and a significant error in detection is marked by a red dotted rectangle}\label{fig5}
\end{figure*}

The instance segmentation results of these methods applied on the test set were visualized (Fig. \ref{fig5}). The HoughNet showed a great error for the sow and even wrongly classified the piglet pixels with the bounding box of the sow as a part of the sow (Figs. \ref{fig5}a1, \ref{fig5}a2, and \ref{fig5}a5). Similarly, the Mask R-CNN also did not accurately detect the large sow object, as the method misclassified farrowing crates that occluded the sow as a part of the sow (Figs. \ref{fig5}b1-b4). YOLACT++ undetected some piglets in two cases (Figs. \ref{fig5}f2 and \ref{fig5}f5). SOLOv2, Blendmask, and CondInst all showed double detection of a single piglet in Fig. \ref{fig5}c1, Figs. \ref{fig5}d2 and \ref{fig5}d4-\ref{fig5}d6, and Figs. \ref{fig5}e2-\ref{fig5}e6, respectively. This happened especially when a piglet was separated into different parts for Blendmask (Fig. \ref{fig5}d6) and CondInst (Fig. \ref{fig5}e6). CClusnet-Inseg demonstrated better instance segmentation results than other methods. Even in the space between two farrowing crates, our method correctly detected the piglets (Fig. \ref{fig5}g3) and the sow (Fig. \ref{fig5}g2).

\subsection{Extension to other applications}
\begin{figure*}[h]%
\centering
\includegraphics[width=0.95\textwidth]{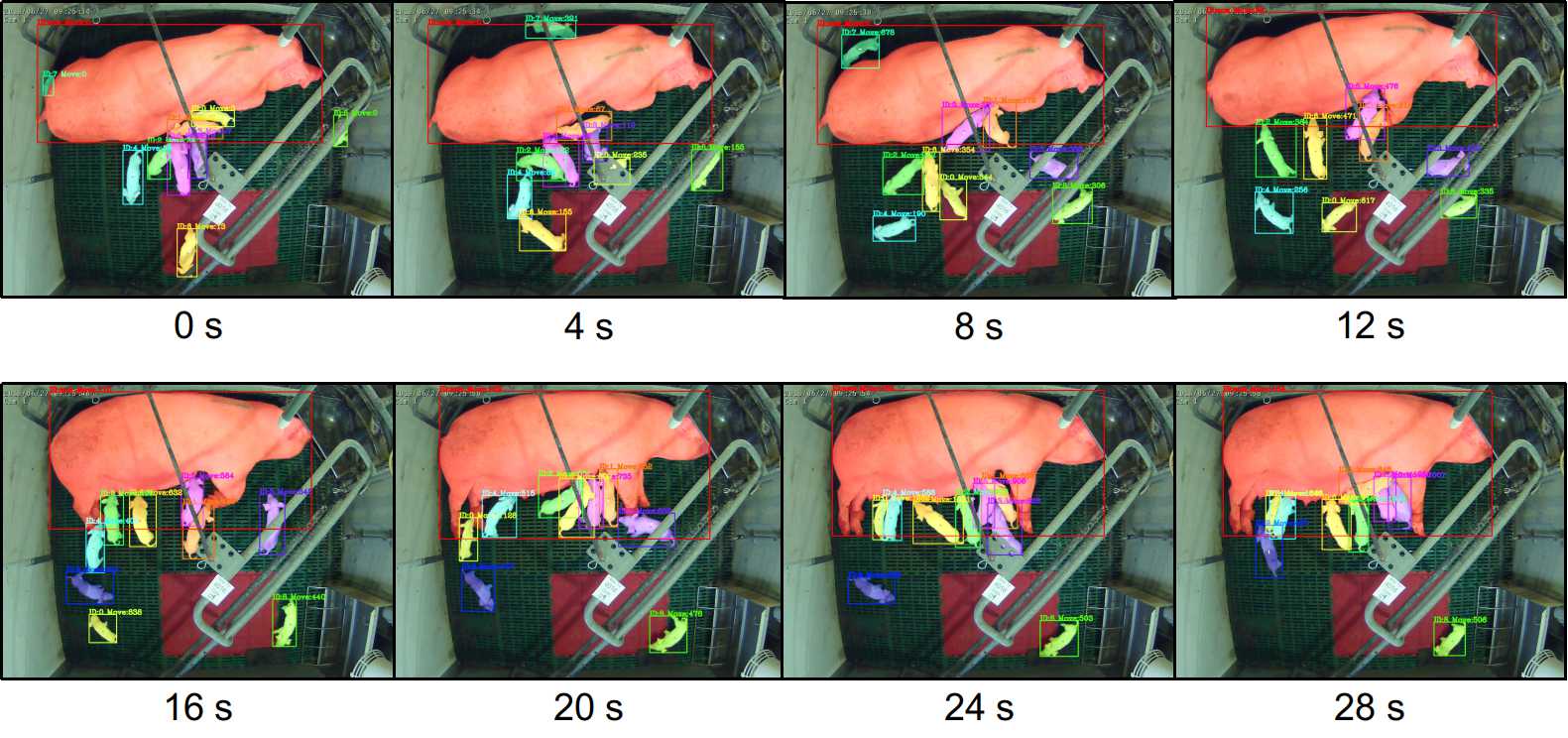}
\caption{Screenshots of multi-object tracking at various time instants, using the results from CClusnet-Inseg}\label{fig6}
\end{figure*}

\begin{figure}[h]%
\centering
\includegraphics[width=0.48\textwidth]{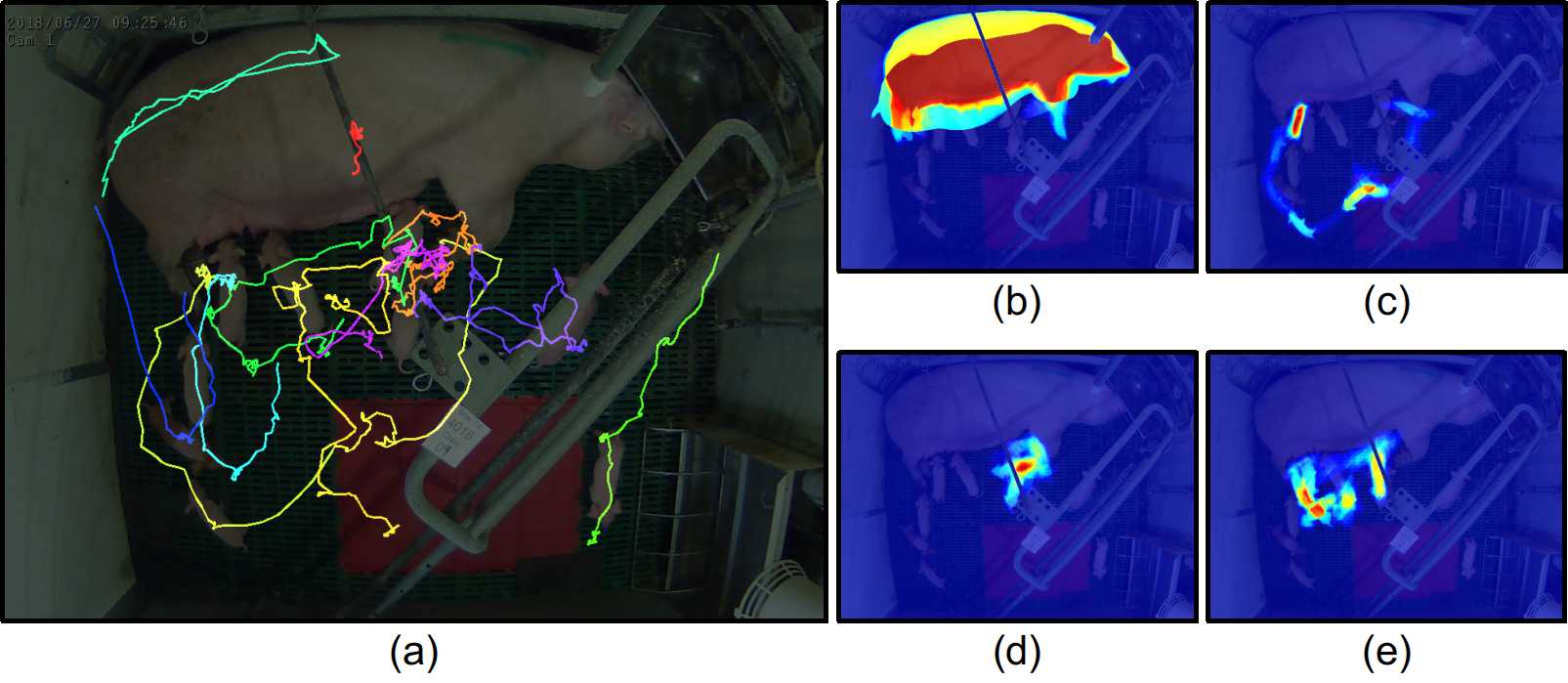}
\caption{Trajectory (a) and spatial distribution of the sow (b) and three piglets (c)–(e). Three piglets with ID = 0, 1, and 2 in the video were selected as an example}\label{fig7}
\end{figure}

To demonstrate the potential applications of our method in animal monitoring, the instance segmentation results were directly applied to multi-object tracking. The video file is available in the Appendix and at \url{https://www.samcityu.com/insegcclusnet}. Screenshots from the video are shown in Fig. \ref{fig6}. From the video, it can be observed that due to its pixel-level accuracy, instance segmentation with IoU showed a small number of errors when pairing objects in consecutive frames, as long as the instance segmentation results were correct. The results of multi-object tracking produced six additional animal characterizations, i.e., trajectory (Fig. \ref{fig7}a), spatial distribution (Figs. \ref{fig7}b–e), movement, speed, body size, and space usage (Table \ref{tab4}) for individual objects. 

The trajectory and spatial distribution with spatial usage could be used to estimate animal comfort, e.g., to determine whether sows make full use of the extra space resulting from opening the crates, and to determine whether the heat pad is effective based on its usage by the piglets (Fig. \ref{fig2}). The movement or average speed of an individual piglet could serve as a piglet vitality index and could be used as a warning of piglet mortality if a piglet does not show movement for a long period. This monitoring is especially important for piglets in the pre-weaning phase. The body pixel size could be used as an interim index for body size estimation and, thus, for body weight and growth estimation; however, the relationship needs further investigation.

\begin{table*}[h]
\begin{center}
\begin{minipage}{\textwidth}
\caption{Results of monitoring through multi-object tracking. The data for three piglets with ID = 0, 1, and 2 in the video are shown as a sample}\label{tab4}
\begin{tabular*}{\textwidth}{@{\extracolsep{\fill}}lcccc@{\extracolsep{\fill}}}
\toprule
Object & Movement (pixel) & Average Speed (pixel/s) & Body Pixel Size (pixel) &Space Usage \\
\midrule
Sow & 155 & 5.4 & 158,341 & 26.9\% \\
Piglet (ID: 0) & 1,383 & 47.9 & 3,375 & 7.3\% \\
Piglet (ID: 1) & 893 & 30.9 & 4,880 & 3.7\% \\
Piglet (ID: 2) & 1,044 & 36.2 & 6,042 & 7.7\% \\
\botrule
\end{tabular*}
\end{minipage}
\end{center}
\end{table*}

Our proposed method is not an end-to-end method and consists of two stages to achieve instance segmentation. Consequently, one of the most critical limitations of our method is the low speed of instance segmentation in comparison with other methods (e.g., Mask R-CNN and YOLACT++). A faster algorithm or even a real-time algorithm is necessary for practical applications; therefore, additional efforts are required for improving the speed, e.g., exploring ways to directly generate the object center as output instead of using a clustering algorithm. 

For generalization, our proposed method uses all visible pixels to detect a target, thus a high utilization of the available information in an image. This high utilization enables our method to manage heavy occlusion cases even though a target is largely occluded. In addition, our method is a target-sensitive and non-occlusion-specific method, since we only detect the pixels belonging to a target and classify all non-target pixels as background. As a result, our method is only sensitive to target pixels, and what the occluder is, what shape it is, and how many parts a target is divided into have few effects on our method. This increases the generalization of our method to be applied to other scenes with different occlusions. Furthermore, our proposed framework can be further expanded using other modules. For example, the encoder-decoder networks can use light networks, including MobileNet \citep{howard2017mobilenets}, EfficientNet \citep{tan2019efficientnet}, and GhostNet \citep{han2020ghostnet}; the clustering algorithm can also be replaced by other algorithms including K-Means and Gaussian Mixed Model. The multiple selections in both two stages enhance the flexibility of our approaches between speed and accuracy, and thus enhance the capacity to fit different animal applications. 

\begin{figure}[h]%
\centering
\includegraphics[width=0.48\textwidth]{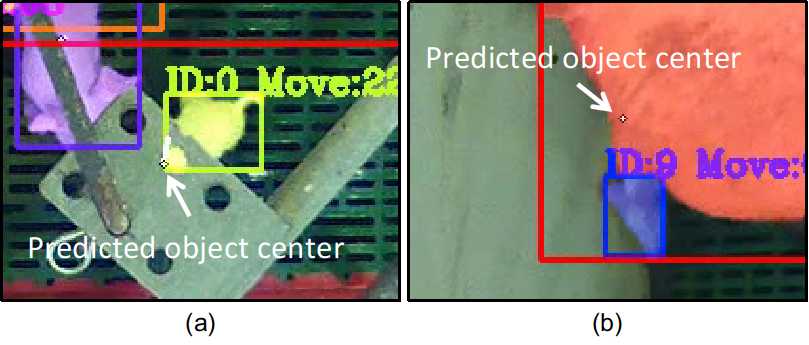}
\caption{Predicted object center using CClusnet-Inseg when the object is under occlusion. (a) A piglet is occluded by farrowing crates. (b) A piglet is occluded by a sow}\label{fig8}
\end{figure}

The object center detected and predicted through our method serves an accurate representation of the object's location, especially when an object is occluded. When occlusions occur, general instance segmentation can identify only the visible pixels of an occluded piglet, and therefore, the location of the piglet (i.e., the center of the physical geometry of the piglet) cannot be accurately represented (Fig. \ref{fig8}). Our method was able to generate an occlusion-resistant object location (using predicted object centers) and, thus, occlusion-resistant movement and trajectory. Thus, our method is advantageous in applications related to animal monitoring, especially for scenes with heavy occlusion. Nevertheless, the detection and prediction of object centers increase the workload related to data labeling in our method. 

We presented only a simple video as an example. Future studies could explore the numerous possibilities in multi-object tracking, e.g., use of the Hungarian algorithm \citep{kuhn1955hungarian} for object pairing and Kalman filter \citep{kalman1960new} for trajectory prediction. In future work, we will explore robust multi-object tracking, and systematic animal monitoring and analysis.

\section{Conclusion}
We developed CClusnet-Inseg, a deep-learning-based instance segmentation method that was specifically designed for pigs in farrowing pens. The main extensions—the DBSCAN algorithm, C2M, and RC2M—were described. The results of an ablation study showed that the model using DBSCAN achieved approximately the same accuracy as and 20 times the speed of the model using mean-shift;  and RC2M improved the mAP of instance segmentation by 1.3 mAP but increased the computation cost by 2.2\%. CClusnet-Inseg achieved an mAP of 84.1 and outperformed other state-of-the-art methods. Our method could be applied in multi-object tracking to generate several animal characterizations (i.e., animal movement, trajectory, average speed, body pixel size, space usage, and spatial distribution), and the predicted object center that is a conjunct output could serve as an occlusion-resistant representation of the object location. 

\bmhead{Acknowledgments}
Thanks to the staff at the swine teaching and research center, School of Veterinary Medicine, University of Pennsylvania for animal care. Funding was provided in part by the National Pork Board and the Pennsylvania Pork Producers Council in the U.S., and the new research initiatives at City University of Hong Kong (Project number: 9610450), Hong Kong SAR, China.

\section*{Statements and Declarations}
\subsection*{Competing Interests and Funding}
The authors have no known competing financial interests or personal relationships that could have appeared to influence the work reported in this paper.

Funding was provided in part by the National Pork Board and the Pennsylvania Pork Producers Council in the U.S., and the new research initiatives at City University of Hong Kong (Project number: 9610450), Hong Kong SAR, China.

\subsection*{Animal ethics statement}
The datasets were generated from animal videos collected in one of our previous studies which was approved by the University of Pennsylvania’s Institutional Animal Care and Use Committee (Protocol \#804656).









\bibliography{sn-article}


\end{document}